\documentclass{article}

\usepackage{spconf,amsmath,graphicx}
\usepackage{booktabs}
\usepackage{amssymb}
\usepackage{multirow}
\usepackage{hyperref}
\usepackage{setspace}
\usepackage{enumitem}

\newcommand{\paragraphHdTop}[1] {\textbf{#1}}

\usepackage[
backend=biber,
style=ieee,
citestyle=numeric-comp,
maxbibnames=3,
maxcitenames=3,
doi=false,isbn=false,url=false,eprint=false
]{biblatex} 

\defbibheading{bibliography}[\refname]{}

\DeclareSourcemap{
	\maps[datatype=bibtex, overwrite=true]{
		\map{
			\step[fieldsource=booktitle,
			match=\regexp{.*Interspeech.*},
			replace={Proc. Interspeech}]
			\step[fieldsource=journal,
			match=\regexp{.*INTERSPEECH.*},
			replace={Proc. Interspeech}]
			\step[fieldsource=booktitle,
			match=\regexp{.*ICASSP.*},
			replace={Proc. ICASSP}]
			\step[fieldsource=booktitle,
			match=\regexp{.*icassp_inpress.*},
			replace={Proc. ICASSP (in press)}]
			\step[fieldsource=booktitle,
			match=\regexp{.*Acoustics,.*Speech.*and.*Signal.*Processing.*},
			replace={Proc. ICASSP}]
			\step[fieldsource=booktitle,
			match=\regexp{.*International.*Conference.*on.*Learning.*Representations.*},
			replace={Proc. ICLR}]
			\step[fieldsource=booktitle,
			match=\regexp{.*International.*Conference.*on.*Computational.*Linguistics.*},
			replace={Proc. COLING}]
			\step[fieldsource=booktitle,
			match=\regexp{.*SIGdial.*Meeting.*on.*Discourse.*and.*Dialogue.*},
			replace={Proc. SIGDIAL}]
			\step[fieldsource=booktitle,
			match=\regexp{.*International.*Conference.*on.*Machine.*Learning.*},
			replace={Proc. ICML}]
			\step[fieldsource=booktitle,
			match=\regexp{.*North.*American.*Chapter.*of.*the.*Association.*for.*Computational.*Linguistics:.*Human.*Language.*Technologies.*},
			replace={Proc. NAACL}]
			\step[fieldsource=booktitle,
			match=\regexp{.*Empirical.*Methods.*in.*Natural.*Language.*Processing.*},
			replace={Proc. EMNLP}]
			\step[fieldsource=booktitle,
			match=\regexp{.*Association.*for.*Computational.*Linguistics.*},
			replace={Proc. ACL}]
			\step[fieldsource=booktitle,
			match=\regexp{.*Automatic.*Speech.*Recognition.*and.*Understanding.*},
			replace={Proc. ASRU}]
			\step[fieldsource=booktitle,
			match=\regexp{.*Spoken.*Language.*Technology.*},
			replace={Proc. SLT}]
			\step[fieldsource=booktitle,
			match=\regexp{.*Speech.*Synthesis.*Workshop.*},
			replace={Proc. SSW}]
			\step[fieldsource=booktitle,
			match=\regexp{.*workshop.*on.*speech.*synthesis.*},
			replace={Proc. SSW}]
			\step[fieldsource=booktitle,
			match=\regexp{.*Advances.*in.*neural.*information.*processing.*},
			replace={Proc. NeurIPS}]
			\step[fieldsource=booktitle,
			match=\regexp{.*Advances.*in.*Neural.*Information.*Processing.*},
			replace={Proc. NeurIPS}]
			\step[fieldsource=booktitle,
			match=\regexp{.*Workshop.*on.* Applications.* of.* Signal.*Processing.*to.*Audio.*and.*Acoustics.*},
			replace={Proc. WASPAA}]
			\step[fieldsource=publisher,
			match=\regexp{.+},
			replace={{}}]
			\step[fieldsource=month,
			match=\regexp{.+},
			replace={{}}]
			\step[fieldsource=location,
			match=\regexp{.+},
			replace={{}}]
			\step[fieldsource=address,
			match=\regexp{.+},
			replace={{}}]
			\step[fieldsource=organization,
			match=\regexp{.+},
			replace={{}}]
		}
	}
}

\title{ESPnet-SLU: Advancing Spoken Language Understanding through ESPnet}
\name{
\begin{tabular}{c}
\it Siddhant Arora${}^1$, Siddharth Dalmia${}^1$, Pavel Denisov${}^2$, Xuankai Chang${}^1$, Yushi Ueda${}^1$,\\ 
\it Yifan Peng${}^1$, Yuekai Zhang${}^3$, Sujay Kumar${}^1$, Karthik Ganesan${}^1$, Brian Yan${}^1$, \\
\it Ngoc Thang Vu${}^2$, Alan W Black${}^1$, Shinji Watanabe${}^1$
\end{tabular}
\vspace{-0.5em}
}
\address{${}^1$Carnegie Mellon University, ${}^2$University of Stuttgart, ${}^3$Zoom Video Communications}

\begin{document}
\ninept
\maketitle
\begin{abstract}
    As Automatic Speech Processing (ASR) systems are getting better, there is an increasing interest of using the ASR output to do downstream Natural Language Processing (NLP) tasks. However, there are few open source toolkits that can be used to generate reproducible results on different Spoken Language Understanding (SLU) benchmarks. Hence, there is a need to build an open source standard that can be used to have a faster start into SLU research. We present ESPnet-SLU, which is designed for quick development of spoken language understanding in a single framework. ESPnet-SLU is a project inside end-to-end speech processing toolkit, ESPnet, which is a widely used open-source standard for various speech processing tasks like ASR, Text to Speech (TTS) and Speech Translation (ST). We enhance the toolkit to provide implementations for various SLU benchmarks that enable researchers to seamlessly mix-and-match different ASR and NLU models. We also provide pretrained models with intensively tuned hyper-parameters that can match or even outperform the current state-of-the-art performances. The toolkit is publicly available at
    https://github.com/espnet/espnet.
\end{abstract}
\begin{keywords}
open-source, spoken language understanding
\end{keywords}
\section{Introduction}
Spoken Language Understanding (SLU) is the task of inferring the semantic meaning of spoken utterances. SLU is an essential component of voice assistants, social bots, and intelligent home devices \cite{socialbot, snips-voice-platform} which have to map speech signals to executable commands every day. Recent advances have driven the commercial success of voice assistants including but not limited to Alexa, Google Home, Siri and Cortana. SLU comprises widespread applications of semantic understanding from spoken utterances. Some examples include recognizing the intent \cite{Lugosch_FSC,SLURP} and their associated entities \cite{SLURP, earnings21} of a user’s command to take appropriate action, or even understanding the emotion behind a particular utterance \cite{IEMOCAP}, and engaging in conversations with a user by modeling the topic of a conversation \cite{SWB_DA_res,SWB}.

Conventional SLU systems consist of a pipeline approach, where a Speech Recognition (ASR) system first maps a spoken utterance into an intermediate text representation, followed by the Natural Language Understanding (NLU) module that extracts the intent from the text representation. Recently, many end-to-end (E2E) SLU \cite{agrawal2020tie,saxon21_interspeech,ganhotra21_interspeech} approaches have been introduced to avoid the error propagation seen in cascaded models. Moreover, these models typically have a smaller carbon footprint \cite{agrawal2020tie} compared to the pipeline-based approaches, making them of particular interest to perform SLU on devices. E2E architectures can also capture non-phonemic speech signals such as pauses, phrasing of words, and intonation which can help provide additional cues towards the semantics that a text-based system cannot capture. These models are also useful for low resource languages \cite{Grabo_data} where there is not enough training data or access to reliable transcripts to separately train ASR and NLU components. 
% These settings are commonly found for non-English languages\cite{Grabo_data,CATSLU} where the industry is more interested in direct understanding than recognition to provide services to speakers of these low resource languages.

With the increase in SLU datasets and methodologies proposed \cite{Lugosch_FSC,coucke2018snips,agrawal2020tie}, there is a growing need for an open-source SLU toolkit which would help standardize the pipelines involved in building an SLU model like data preparation, model training, and its evaluation. Our goal is to provide an open-source standard where researchers can easily incorporate previously proposed technologies, compare and contrast new ideas with the existing methodologies. In this work, we introduce a new E2E-SLU toolkit built on an already existing open-source speech processing toolkit ESPnet \cite{ESPnet,ESPnet-TTS,ESPnet-ST}. ESPnet supports a variety of speech processing tasks ranging from front-end processing like enhancement and separation to recognition and translation. Having ESPnet-SLU would help users build systems for real-world scenarios where many speech processing steps need to be applied before running the downstream task. ESPnet also provides an easy access to other speech technologies being developed like data-augmentation \cite{SpecAugment}, encoder sub-sampling \cite{ESPnet}, and speech-focused encoders like conformers \cite{Conformer}. They also support many pretrained ASR \cite{Hubert,Wav2vec2,TERA,VQ-APC} and NLU systems \cite{BERT,MPNet} that can be used as feature extractors in a SLU framework. 

\noindent The contributions of ESPnet-SLU are summarized below:
\begin{table*}
\caption{Comparison with other open-source End to End Spoken Language Understanding toolkits in September 2021
}
\label{tbl:espnet_comp}
\centering
 \resizebox {0.9\linewidth} {!} {
\begin{tabular}{l|ccccc}
\toprule
 & Alexa\cite{agrawal2020tie} & Lugosch\cite{Lugosch_FSC} & CoraJung \cite{cha2021speak} & SpeechBrain\cite{SpeechBrain} &  \bf{ESPnet-SLU} \\\midrule
BiLSTM based encoder & \checkmark & \checkmark & \checkmark & \checkmark & \checkmark \\
Transformer based encoder & & & & \checkmark & \checkmark \\
Conformer based encoder &  & & & \checkmark & \checkmark \\                   \midrule
Classifier & \checkmark &  & & \checkmark & \\
RNN based decoder &  & \checkmark & \checkmark & \checkmark & \checkmark \\
Transformer based decoder & & & & \checkmark  & \checkmark \\ \midrule
Supports multi tasking with ASR? &  & & & & \checkmark \\
Supports multi tasking with NLU? & \checkmark & & \checkmark & &  \\
Supports using pretrained ASR model? &  & \checkmark & \checkmark & \checkmark & \checkmark  \\
Supports using pretrained NLU model? & \checkmark & & \checkmark & & \checkmark \\
% Supports using data augmentation? & & & \checkmark & & \checkmark \\
Supports other task? &   &  & & \checkmark & \checkmark \\
Supports SLU on languages besides English? &   &  & & & \checkmark \\
Supports using context from previous utterances? &   &  & & & \checkmark \\
Supports using tasks in pipeline manner? &  &  & & & \checkmark \\
Provide pretrained model &  & \checkmark & & \checkmark  & \checkmark \\
\bottomrule
\end{tabular}
}
\vspace{-10px}
\end{table*}
% \begin{itemize}
\begin{itemize}[leftmargin=*,itemsep=0pt, topsep=1pt]
    \item We provide recipes that covers all experiment processes for intent classification \cite{Lugosch_FSC,coucke2018snips}, slot filling \cite{SLURP}, emotion recognition \cite{IEMOCAP} and dialogue acts classification \cite{SWB} datasets. The toolkit also contains implementations in non English languages \cite{Grabo_data,CATSLU,yoshino18_LREC,karunanayake-etal-2019-transfer}.
    \item This toolkit incorporates the use of pretrained ASR  models like HuBERT, Wav2vec2 and pretrained NLU models like BERT, MPNet that can be used as feature extractors for ASR and NLU submodules inside the E2E-SLU framework.
    \item It also contains implementations of various speech processing tasks that can be used in a pipeline manner, thus replicating real-world scenarios where speech processing frontend need to be applied before performing a downstream task\footnote{The interactive demo on - \url{https://espnet-slu.github.io}}.%colab.research.google.com/drive/14nCrJ05vJcQX0cJuXjbMVFWUHJ3Wfb6N?usp=sharing}}.
    \item We release an open-source toolkit and provides easy access to the trained models that match or even significantly outperform the state-of-the-art performance on these benchmarks.
\end{itemize}

\section{Design}
This section briefly describes the design for recipes that include all procedures to complete model training and evaluation on a given dataset. The recipes have been carefully designed to follow a unified approach with stage-by-stage processing as described in \cite{ESPnet-SE}. Table~\ref{tbl:espnet_comp} summarises the features supported by our toolkit and other popular E2E-SLU toolkits to the best of our knowledge. 
We compare with 4 well maintained frameworks, i.e. Alexa (alexa-end-to-end-slu)~\cite{agrawal2020tie}, Lugosch (lorenlugosch/end-to-end-SLU)~\cite{Lugosch_FSC}, CoraJung (CoraJung/flexible-input-slu)~\cite{cha2021speak} and SpeechBrain~\cite{SpeechBrain}. \footnote{There are many excellent toolkits\cite{ultes2017pydial,uberplato,rasa} that support only NLU task.}
% All the frameworks are based on PyTorch and implement SOTA SLU methods, along with the training and evaluation scripts. The first 3 frameworks do not provide implementation for multiple SLU architectures thus not facilitating model comparison. 

\paragraphHdTop{Recipes}
\label{Recipes}
%  We provide various recipes in order to easily implement a strong baseline across a variety of datasets. We broadly categorise our implementation in 3 types of data regimes; (1) Mid resource datasets, which comprises of majority of the SLU datasets, like Fluent Speech Commands (FSC) \cite{Lugosch_FSC}, and Snips SmartLight (Snips) \cite{coucke2018snips} do not contain sufficient audio files to train the ASR module from scratch, and pretrained ASR model can be effectively utilized to improve acoustic modeling. Further, multi-tasking with ASR transcripts can help further improve model performance. (2) High resource SLU datasets like SLURP \cite{SLURP} that provides both intent and transcript for a large number of audio files. SLU architecture on these datasets usually do not benefit from using pretrained models, and we can achieve high accuracy through better model architectures. (3) Most datasets on i18n (non English) languages can be described as low resource datasets like Grabo \cite{Grabo_data} where the dataset often lacks speech data, and multilingual pretrained ASR models are used as feature extractors. In these scenarios, the transcripts are often not available or reliable to perform ASR multi-tasking. By providing recipes for each of these types of datasets, the toolkit facilitates researchers to understand what methodologies work in different data regimes.
 We provide various recipes in order to implement a strong baseline across a variety of datasets. We broadly categorize our implementation into 3 types of data regimes; (1) Mid resource datasets, which comprises of majority of the SLU datasets like Fluent Speech Commands (FSC) \cite{Lugosch_FSC}, and Snips SmartLight (Snips) \cite{coucke2018snips}. Usually, they do not contain sufficient data to train an ASR module from scratch, but a pretrained ASR model can help improve acoustic modeling. Multi-tasking SLU with ASR transcripts can further improve model performance. (2) High resource SLU datasets like SLURP \cite{SLURP} provide both intent and transcript for a large number of audio files. Models on these datasets can utilize the ASR transcripts for multi-tasking for improved performance, but they do not usually benefit from pretrained models. (3) Most SLU datasets on non-English languages can be described as low resource like the Dutch Grabo dataset \cite{Grabo_data}. They often lack speech data and hence multilingual pretrained ASR models can help as feature extractors. In these scenarios, transcripts are often unavailable or unreliable to perform ASR multi-tasking. By providing recipes for each of these datasets, ESPnet-SLU facilitates researchers to understand what methodologies work in different data regimes.
 
\paragraphHdTop{Tasks}
\label{ESPnet-tasks}
ESPnet supports various speech processing tasks such as ASR \cite{ESPnet}, TTS \cite{ESPnet-TTS}, ST \cite{ESPnet-ST}, SE \cite{ESPnet-SE} and Voice Conversion (VC) \cite{ESPnet-VC}. We believe that to perform downstream understanding tasks on real-world audio, these tasks need to be applied in conjunction with SLU. By having multiple tasks in a single unified implementation, ESPnet allows the use of different speech tasks in a pipeline manner that can have widespread applications, as shown in Section \ref{task-pipeline}.

\paragraphHdTop{ASR Multi-task learning}
\label{ASR-Multitask}
Since SLU requires both acoustic and semantic understanding, it is often regarded as a more challenging task than ASR and NLU. Multi-task learning-based approaches \cite{li-etal-2020-multi,agrawal2020tie,cha2021speak} have become popular to strengthen the training of SLU systems. Hence, we allow the option to add auxiliary ASR objectives by making the model generate both intent and transcript. 
% We are the first popular open-source SLU toolkit to support ASR multi-tasking, as shown in Table \ref{tbl:espnet_comp}.

\paragraphHdTop{ASR and NLU pretraining}
\label{Pretrained-Model}
Recent work \cite{Pretrain_ASR,wang2021pretraining} has advanced the state-of-the-art SLU performance by building the architecture on self-supervised ASR and NLU models. Inspired by this work, we also support options to use our framework's pretrained ASR and NLU models as feature extractors. More details are in section \ref{ref:models}.

\paragraphHdTop{Low resource Multilingual SLU}
\label{Multi-lang}
The toolkit also contains recipes for languages such as Japanese \cite{yoshino18_LREC}, Dutch \cite{Grabo_data}, Tamil \cite{karunanayake-etal-2019-transfer}, Sinhala\cite{karunanayake-etal-2019-transfer} and Mandarin \cite{CATSLU}. With these recipes, we want to facilitate research in SLU technologies and ensure that they are available to a wide variety of users, going beyond English-speaking users.

\paragraphHdTop{Combining context from previous utterances}
\label{Combine-context}
Human interactions are usually in the form of spoken conversations, where the semantic meaning of a given utterance depends on the context \cite{kim-etal-2019-gated, kim19cross, ganhotra21_interspeech} in which it was spoken. Hence, we support using dialogue history to perform classification on each conversation turn.
\begin{table}
\caption{Supported tasks and datasets in ESPnet-SLU along with their reported performance in the original paper and our toolkit. We show the metrics used in the original paper. We match or outperform SOTA performance across a variety of SLU benchmarks.}
\centering
 \resizebox {\linewidth} {!} {
\begin{tabular}{c|llcc}
\toprule
 Task & Dataset & Metric & Paper Results & \bf{ESPnet-SLU} \\\midrule
\multirow{11}{*}{IC} & SLURP \cite{SLURP} & Acc. & 78.3 & 86.3 \\
 & FSC \cite{Lugosch_FSC} & F1 & 98.8 & 99.6 \\
 & FSC Unseen (S) \cite{Lugosch_FSC,FSC_MASE} & Acc. & 94.2 & 98.6 \\
 & FSC Unseen (U) \cite{Lugosch_FSC,FSC_MASE} & Acc. & 88.3 & 86.4 \\
 & FSC Challenge (S) \cite{Lugosch_FSC,FSC_MASE} & Acc. & 92.3 & 97.5 \\
 & FSC Challenge (U) \cite{Lugosch_FSC,FSC_MASE} & Acc. & 78.3 & 78.5 \\
 & SNIPS \cite{coucke2018snips} & F1 & 91.7 & 91.7 \\
 & HarperValleyBank \cite{Harper_Valley} & Acc & 45.5 & 47.1\\
 & Grabo \cite{Grabo_data,Grabo_res} & Acc. & 94.5 & 97.2\\ 
 & CAT-SLU MAP \cite{CATSLU,CATSLU_data} & Acc. & 79.8 & 78.9 \\
 & Speech Commands \cite{warden2018speech} & Acc. & 88.2 & 98.4 \\ \midrule
 SF & SLURP \cite{SLURP} & SLU-F1 & 70.8 & 71.9\\\midrule
 \multirow{2}{*}{DA} & Switchboard \cite{godfrey1992switchboard,jurafsky1997switchboard}  & Acc. & 68.7 & 67.5\\ 
 & HarperValleyBank \cite{Harper_Valley}  & Acc. & 45.5& 47.1\\ \midrule
 ER & IEMOCAP \cite{IEMOCAP,yang2021superb}  & 5-fold Acc. & 67.6 & 69.4 \\
\bottomrule
\end{tabular}
}
\label{tbl:espnet_tasks}
\vspace{-10px}
\end{table}

\section{Example Models}

\label{ref:models}
To provide a glimpse into various models supported within our SLU toolkit, we briefly describe the construction of an example E2E-SLU model. The library is written in python using PyTorch as the main neural network library. The following sections describe the general details without going into the dataset-specific customizations.

\paragraphHdTop{Encoder Decoder Model}
We build the SLU model as a Transformer-based hybrid CTC/attention framework \cite{HybridCTC}. The transformer architecture \cite{Transformer_ASR} usually consists of 12 self-attention blocks in the transformer encoder and 6 self-attention blocks in the decoder. We also experiment with Conformer \cite{Conformer} encoders. 

\paragraphHdTop{Using pretrained ASR models as pre-encoder}
We support using pretrained ASR models as feature extractors for our encoder architecture. We use the s3prl \cite{yang2021superb} and fairseq \cite{ott2019fairseq} toolkit to access a variety of self-supervised learning representations as frontend in our SLU architecture. These pretrained ASR models are inserted before the Encoder such that Encoder takes in the output of these ASR models as acoustic features extracted from the input audio file.

\paragraphHdTop{Using pretrained NLU models as post-encoder}
We integrate the HuggingFace Transformers library \cite{wolf-etal-2020-transformers}, which allows usage of numerous generic and task-specific pretrained NLU models.
%, many of which reach SOTA scores on their respective NLU tasks. 
Pretrained self-attention blocks of the NLU model can be inserted into any sequence-to-sequence model between the Encoder and Decoder components and therefore we name this component \emph{post-encoder NLU}. In this configuration, hidden states from Encoder output are passed to the first self-attention block of NLU instead of NLU token embeddings, and Decoder consumes the output of the last NLU self-attention block instead of Encoder output. This way, the output of Encoder gets processed by NLU and may have more information about its linguistic properties, e.g. semantics. 
%Such information flow through differentiable operations in continuous space allows complete end-to-end optimization of SLU performance using standard deep learning methods unlike in the conventional pipeline ASR/NLU approach.
%We also experimented with a post encoder layer that takes hidden representations from the encoder and outputs representations to the decoder. Different pretrained NLU models provided in the Hugging Face repository can be utilized as a post encoder to incorporate semantic information into the framework.

% \paragraphHdTop{Using ASR tasks in pipeline manner}

\section{Experiments}
\begin{table}[]
\caption{Intent Classification accuracy
on FSC \cite{Lugosch_FSC} for models using ASR multitasking, pretrained ASR and data augmentation methods. SpeechBrain \cite{SpeechBrain} results are accessed on September 2021.
}
 \centering
  \resizebox {\linewidth} {!} {
\begin{tabular}{l|l|c}
\toprule
& Model                                                             & IC (\% Acc)  \\\midrule
\multirow{2}{*}{\shortstack[l]{Baseline}}& E2E-SLU \cite{Lugosch_FSC}                                         & 96.6 \\
& + Pretraining ASR  \cite{Lugosch_FSC}                            & 98.8 \\ 
& Pretrained E2E-SLU + data augmentation \cite{SpeechBrain}       & 99.6 \\ \midrule
\multirow{4}{*}{\textbf{ESPnet-SLU}} & Tsf. Encoder w/ Full Intent Decoding & 93.5 \\
& + SpecAug Data Augmentation                         & 98.9 \\ 
& \hphantom{0000}+ ASR Multi-tasking  & 99.4 \\
& \hphantom{000000}+ Pretrained ASR HuBERT & \bf{99.6}\\\midrule
\multirow{3}{*}{\shortstack[l]{Ablations\\for Intent\\Decoding}}
& ESPnet-SLU w/ Character Decoding     & 98.3 \\
& w/ Slot Decoding        & 97.8 \\ 
& w/ Full Intent Decoding     & 98.9 \\
\bottomrule
\end{tabular}
}
\label{tbl:FSC}
\vspace{-10px}
\end{table}

In this section, we demonstrate how models from our toolkit described in Section \ref{ref:models} perform on benchmark spoken language understanding datasets i.e. intent classification (IC) \cite{Lugosch_FSC,SLURP,coucke2018snips,Harper_Valley,warden2018speech}; slot filling (SF) \cite{SLURP}; emotion recognition (ER) \cite{IEMOCAP} and dialogue acts (DA) classification \cite{SWB_DA_res} corpora. As discussed in Section \ref{Recipes}, we also perform experiments on low resource non-English datasets \cite{Grabo_data,CATSLU}. 
% Detailed training and decoding configurations
% are available in conf/train\_asr.yaml and
% conf/decode\_asr.yaml, respectively. 
The detailed comparison with the results in the original dataset paper is shown in Table~\ref{tbl:espnet_tasks}. All the results are reported on the splits provided by the original paper's authors unless stated otherwise.
\begin{table}[]
\caption{Intent Classification F1 score on Snips \cite{coucke2018snips} where we experiment with finetuning the frontend pretrained ASR models. }
\label{tbl:Snips}
 \centering
\resizebox {0.7\linewidth} {!} {
\begin{tabular}{l|c}
\toprule
Model                                                             & IC (F1) \\\midrule
Pipeline ASR + NLU \cite{coucke2018snips}   & \bf{91.7} \\ \midrule
\textbf{ESPnet-SLU} w/ Pretrained HuBERT & 87.4 \\
\hphantom{00} + Finetuning HuBERT     & 89.1 \\
\hphantom{0000} + ASR Multi-tasking     & \bf{91.7} \\
\bottomrule
\end{tabular}
}
\vspace{-10px}
\end{table}
\subsection{Intent Classification (IC) and Slot Filling (SF)}
The intent classification task is modeled as a conditional prediction task where we decode the intent as one word. Slot filling is modeled similarly where we first predict intent followed by entity label and lexical filler, separated by separator tokens.
% We also experiment with training ASR system to infer the written transcript after the intent to perform ASR multitasking along with intent classification. Our analysis also evaluated the use of different pretrained ASR and NLU systems to incorporate additional acoustic and semantic information in our framework.

\paragraphHdTop{FSC (IC)} \cite{Lugosch_FSC} tests a model’s ability to predict intents from commands used with an intelligent home voice assistant. Table \ref{tbl:FSC} shows the result of different model architectures on this benchmark. We observe that a transformer-based SLU system with SpecAug \cite{SpecAugment} data augmentation can outperform the published results achieved by using a pretrained ASR system \cite{Lugosch_FSC} on this dataset. Also, instead of decoding the intent as a whole word, we tried decoding intent by character and by each slot which was found to hurt the intent classification performance.  
We experimented with multitasking with ASR transcripts as discussed in Section \ref{ASR-Multitask}, gaining further improvements. Finally, using the pretrained ASR model HuBERT as a feature extractor improved the acoustic modeling, achieving SOTA performance \cite{SpeechBrain} on this dataset. We also show results on the recently proposed Challenge and Unseen split \cite{FSC_MASE} in Table~\ref{tbl:espnet_tasks} where we are able to outperform baselines in unseen speaker(S) test set and match baselines for unseen utterance(U) test set.

\paragraphHdTop{Snips (IC)} \cite{coucke2018snips} is another popular SLU benchmark whose results are shown in table \ref{tbl:Snips}. We perform our experiments on random split constructed using the approach followed in \cite{agrawal2020tie}. We observe that finetuning pretrained ASR models can further help in improving performance. Unlike FSC, Snips had unseen utterances in the test set that were not observed during training. Hence, we performed byte pair encoding of the transcript before concatenating with the intent to reduce the mismatch in the vocabulary of training and test transcripts and match the baseline performance.

\begin{table}[]
 \caption{Intent Classification accuracy on the SLURP Dataset \cite{SLURP} where we perform comparison between different pretrained ASR and NLU systems as feature extractors. SpeechBrain \cite{SpeechBrain} results are accessed on September 2021.
}
 \centering
  \resizebox {\linewidth} {!} {
\begin{tabular}{l|l|c}
\toprule
& Model                                                             & IC (F1) \\\midrule
\multirow{3}{*}{Baseline}& Pipeline ASR+NLU w/ synthetic data~\cite{SLURP} & 74.6 \\
& \hphantom{000}+ Additional ASR data \cite{SLURP}              & 78.3 \\ 
& E2E-SLU w/ Pretraining + synthetic data \cite{SpeechBrain}       & 75.1 \\ \midrule
% & \bf{ESPnet-SLU} & \\
\multirow{3}{*}{\bf ESPnet-SLU} & E2E-SLU w/ Conformer Encoder & 76.4 \\
& \hphantom{000}+ Pretrained ASR HuBERT \cite{Hubert} & 77.0 \\
& + synthetic data & \bf{86.3} \\ \midrule
\multirow{4}{*}{\shortstack[l]{Ablations for \\ Pretrained ASR}}& \hphantom{000}+ VQ-APC \cite{VQ-APC}  & 82.1 \\
& \hphantom{000}+ HuBERT \cite{Hubert}    & 83.3 \\
& \hphantom{000}+ Wav2vec2 \cite{Wav2vec2}     & 83.3 \\
& \hphantom{000}+ TERA \cite{TERA}   & 83.5 \\
 \midrule
\multirow{2}{*}{\shortstack[l]{Ablations for \\ Pretrained NLU}}& \hphantom{000}+ MPNET \cite{MPNet} & 82.5 \\
& \hphantom{000}+ BERT \cite{BERT} & 85.7 \\
% \midrule
% \hphantom{000}+ Cascade SLU &  \\
% \hphantom{000}+ Multi Decoder SLU &  \\
% Conformer ASR+ Transformer NLU                                  & 66.2 \\
% +Pretrain Hubert                         & 75.7 \\
% + BPE    & 77.6 \\
% + Synethtic Dataset        & 83.3 \\
\bottomrule
\end{tabular}
}
\label{tbl:SLURP}
\vspace{-10px}
\end{table}
\paragraphHdTop{SLURP (IC, SF)} \cite{SLURP} has been recently proposed as a substantially larger and more linguistically diverse SLU dataset. It consists of prompts for an in-home personal robot assistant. Unlike FSC and Snips, pretrained ASR systems did not significantly improve performance on this larger SLU dataset. Including the provided synthetic dataset (SLURP-synth) into our training set, as done in \cite{SLURP}, achieved a significant 8\% performance gain over the previous state-of-the-art on this benchmark which is a pipeline model that uses additional ASR training data. We also analyzed using different pretrained ASR systems as feature extractors and observed that they did not help improve performance over FBANK. Thus, in contrast to results in SUPERB \cite{yang2021superb} benchmark, the pretrained ASR systems do not always improve performance when used as feature extractors in an E2E SLU system. 
% We observe that Masked Reconstruction based pretrained ASR system TERA \cite{TERA} performs best, whereas autoregressive reconstruction based pretrained ASR system VQ-APC \cite{VQ-APC} performs worst among different pretrained ASR systems. 
We also analyzed the impact of pretrained NLU systems to incorporate semantic information. However, we observed no gains in performance. This analysis shows that researchers can use our toolkit to compare the utility of different pretrained ASR and NLU systems as feature extractors (See Section \ref{Pretrained-Model}) for intent classification.
We also perform the Slot Filling (Entity Classification) task on SLURP \cite{SLURP} dataset. 
% It is again modeled as a generative task where we predict the first word as intent followed by entity label and lexical filler separated by separator token and the ASR transcripts. 
As shown in Table \ref{tbl:espnet_tasks}, we outperform the previous best SLU-F1 \cite{SLURP} performance.

\paragraphHdTop{Non English SLU (IC)} i.e., for Dutch (Grabo Dataset \cite{Grabo_data}) and Mandarin (CAT-SLU MAP \cite{CATSLU}). For CAT-SLU, we use multilingual pretrained ASR model XLSR-53 \cite{XLSR-53} as the frontend, whereas we do not use any pretrained ASR models for the Grabo dataset. To simulate a low resource setting discussed in Section \ref{Recipes} in the Grabo dataset, we do not concatenate the transcript with intent and are still able to outperform the no pretrained ASR results reported in \cite{Grabo_res}.

\paragraphHdTop{Other Corpora (IC)} The performance for other intent classification datasets is shown in Table~\ref{tbl:espnet_tasks}, demonstrating the broad coverage of our system. Google Speech Commands \cite{warden2018speech} is a dataset used to train a limited domain ASR system on which we were able to outperform prior best performance. 
We were able to match intent classification results on the HarperValleyBank corpus \cite{Harper_Valley} which is a corpus of spoken dialog between an agent and a customer of a bank.
\begin{table}[t]
\caption{Emotion Recognition accuracy of ESPnet-SLU models on the IEMOCAP Dataset \cite{IEMOCAP} with different pretrained ASR systems. We report results on 1 out of 5 folds for development. SpeechBrain \cite{SpeechBrain} results are accessed on September 2021.}
 \centering
\resizebox {0.95\linewidth} {!} {
\begin{tabular}{l|c}
\toprule
Model & ER (\% Acc)  \\ \midrule
E2E-SLU \cite{SpeechBrain}       & 65.7 \\ \midrule
\textbf{ESPnet-SLU} w/ Conformer Enc. + ASR Multi-task & 57.5  \\ 
% +Concatenate transcript & 57.5\\ \midrule
+ Pretrained ASR Wav2vec2 \cite{Wav2vec2} & 67.6\\
+ Pretrained ASR HuBERT \cite{Hubert} & \bf{70.0} \\
\bottomrule
\end{tabular}
}
\label{tbl:IEMOCAP}
\vspace{-10px}
\end{table}
\begin{table}[t]
\caption{Dialogue Act Classification accuracy results on the SWB Dataset \cite{SWB_DA_res} showing the impact of using spoken dialog contexts. }
\label{tbl:SWB_DA}
 \centering
\resizebox {0.9\linewidth} {!}  {
\begin{tabular}{l|c}
\toprule
Model                & DA (\% Acc)  \\\midrule
Pretrained ASR + NLU w/ 2 utt. context \cite{ortega2019context} &  \bf{68.7} \\
Baseline E2E-SLU \cite{SWB_DA_res}        & 50.9 \\ 
\midrule
\bf{ESPnet-SLU} w/ Conformer & 52.9 \\
\hphantom{0} + 3 utterance context &  54.9 \\
\hphantom{000} + Pretrained ASR HuBERT \cite{Hubert}  & 67.5 \\
\bottomrule
\end{tabular}
}
\vspace{-10px}
\end{table}
\vspace{-10px}
\subsection{Emotion Recognition (ER)}
Emotion Recognition is also modeled as a conditional prediction task where we infer the first word as the emotion class. We conduct our experiments on IEMOCAP \cite{IEMOCAP} dataset using the four classes (neutral, happy, sad, angry). Model comparisons were made based on the split using Sessions 1--4 as a training set and Session 5 as a test set. As seen in Table~\ref{tbl:IEMOCAP}, using a pretrained HuBERT model before the conformer encoder performs the best. Next, we compare the accuracy of this model with the one reported in \cite{yang2021superb} based on 5-fold cross-validation in Table \ref{tbl:espnet_tasks}, and achieve competitive performance.

\subsection{Dialogue Act Classification (DA)}
Dialogue Act classification is modeled similar to the intent classification task. Given an utterance, the system has to classify it to one of the DA classes, such as statement, question, etc. We conduct our experiments on  NXT-format Switchboard Corpus that annotates Switchboard telephone speech corpus \cite{godfrey1992switchboard} with 42 DA classes \cite{jurafsky1997switchboard}. Since it has been reported that context is important for DA classification \cite{lee2016sequential,ortega2017neural}, we also extend each utterance by simple concatenation with the acoustic signal from 3 preceding utterances to provide context (see Section ~\ref{Combine-context}) which improves the accuracy by 2\%. Furthermore, using pretrained HuBERT ASR models increases the accuracy to 67.7\%, which is close to baseline accuracy on this dataset.
\begin{figure}[t]
  \centering
    \includegraphics[width=0.9\linewidth]{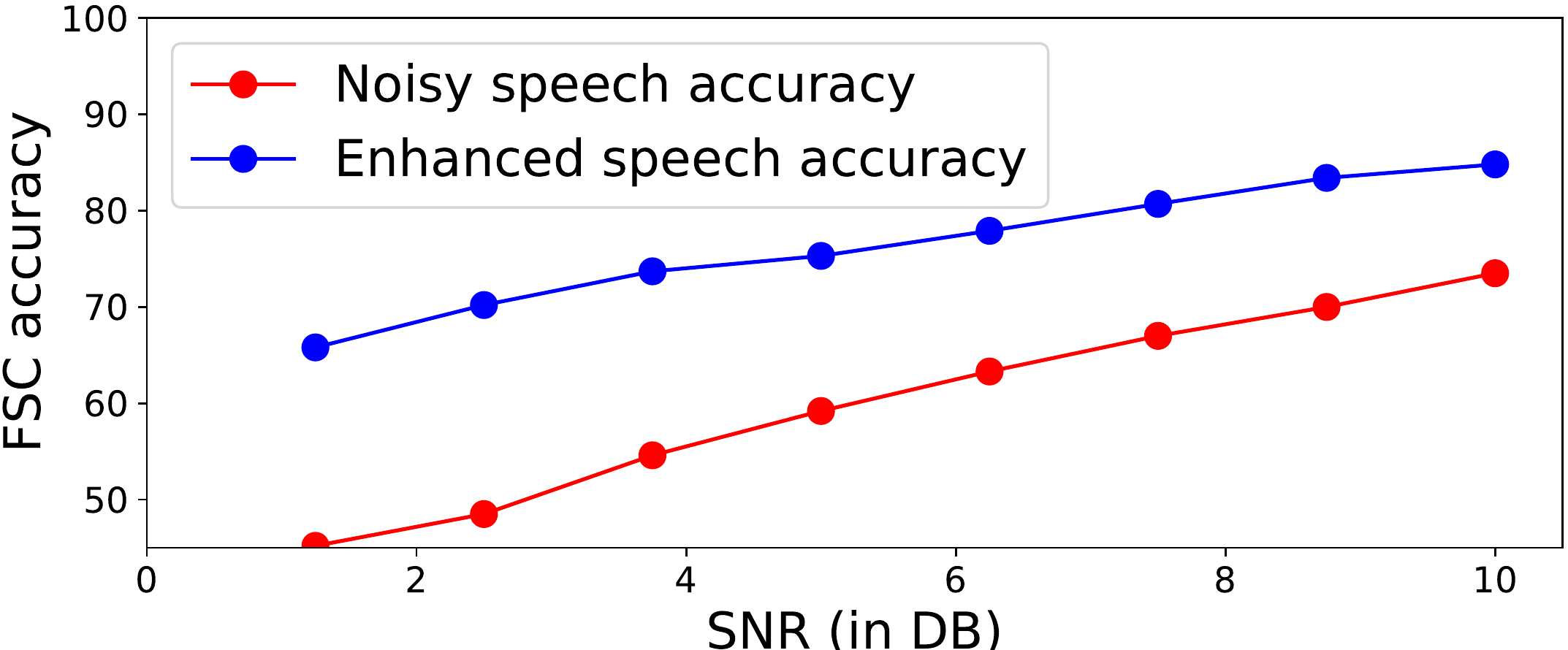}
    \vspace{-10pt}
  \caption{Intent classification accuracy on the FSC dataset against the Signal-to-Noise Ratio (SNR) of noisy speech. This plot indicates that applying Speech Enhancement (SE) before running our SLU model reduces the performance drop with no-noise speech.}
  \label{img:noisy_FSC}
\end{figure}

\subsection{Noisy Intent Classification (IC) with Speech Enhancement}
\label{task-pipeline}
As discussed in Section \ref{ESPnet-tasks}, ESPnet already has the implementation of various speech processing tasks like ASR, SE, and many more. This experiment explores the effectiveness of supporting numerous tasks in a single toolkit by using multiple ASR tasks in a pipeline manner. We test our hypothesis on the Fluent Speech Command dataset. We first convert the audio files into noisy speech by adding real-world noise \cite{Wichern2019WHAM}. We computed the intent classification performance using our already trained model on clean audio files and observed a significant drop in performance in Figure~\ref{img:noisy_FSC}. The noise files were then passed through a speech enhancement model trained on CHIME4 \cite{CHIME4} dataset. We observe a significant improvement in intent classification performance on these enhanced audio files, highlighting the advantage of having multiple tasks in a unified toolkit. 
% Our observations were consistent for different values of SNR.
% \input{tables/compare_pretrain_ASR.tex}

\section{Conclusion}
% We present ESPnet-SLU, a new E2E-SLU toolkit, which is an extension of the popular open source toolkit ESPnet. 
% It facilitates the fast open-source development of SLU systems by standardizing recipes for various benchmarks containing data preparation, training, and model evaluation. 
We present ESPnet-SLU, a new open-source E2E-SLU toolkit, with the objective of facilitating fast research and development of SLU systems through standardized recipes for various benchmarks containing data preparation, training, and model evaluation. 
ESPnet-SLU contains recipes for over 10 diverse SLU corpora, encompassing multiple languages and task types, with performance nearing or exceeding the prior state-of-the-art. Furthermore, our design is a modular extension of the popular ESPnet toolkit with access to the entire pre-existing infrastructure of various speech processing tasks, models and architectures. 
% We hope that this work stimulates further research in SLU. 
In the future, we will support more corpora and implement more SLU systems like NLU multi-tasking to further advance the performance of our SLU systems.

% A core feature that differentiates it from others is ASR multi-tasking and providing support of SLU systems in multiple languages.

\section{Acknowledgements}
This work used the Extreme Science and Engineering Discovery Environment (XSEDE), which is supported by NSF grant number ACI-1548562.
Specifically, it used the Bridges system, supported by NSF grant ACI-1445606, at the PSC.

% Further, by extending over the ESPnet toolkit ESPnet-SLU has access to the entire infrastructure various speech processing tasks, models and architectures supported by ESPnet.

% \vfill\pagebreak

\label{sec:refs}

% \bibliographystyle{IEEEbib}
% \bibliography{strings,refs}
\section{References}
{
\setstretch{0.85}
\printbibliography
}
\end{document}